\documentclass{article} %

\usepackage[accepted]{icml2018}

\usepackage{amsmath,amsfonts,bm}

\def\eqref#1{equation~\ref{#1}}

\def\1{\bm{1}}

\DeclareMathAlphabet{\mathsfit}{\encodingdefault}{\sfdefault}{m}{sl}
\SetMathAlphabet{\mathsfit}{bold}{\encodingdefault}{\sfdefault}{bx}{n}

\usepackage{microtype}
\usepackage{subfigure}
\usepackage{booktabs} %
\usepackage{multirow}

\usepackage{hyperref}
\usepackage{url}
\usepackage[utf8]{inputenc}
\usepackage[disable,textsize=tiny,textwidth=8em,backgroundcolor=yellow]{todonotes}

\usepackage{natbib}
\usepackage{graphicx}

\usepackage[title]{appendix}

\begin{document}

\twocolumn[
\icmltitle{Analyzing and Improving Representations with the Soft Nearest Neighbor Loss}

\icmlsetsymbol{equal}{*}

\begin{icmlauthorlist}
\icmlauthor{Nicholas Frosst}{brain}
\icmlauthor{Nicolas Papernot}{brain}
\icmlauthor{Geoffrey Hinton}{brain}
\end{icmlauthorlist}

\icmlaffiliation{brain}{Google Brain}

\icmlcorrespondingauthor{N. Frosst}{frosst@google.com}
\icmlcorrespondingauthor{N. Papernot}{papernot@google.com}

\icmlkeywords{Machine Learning, ICML}

\vskip 0.3in
]

\printAffiliationsAndNotice{}  %

\begin{abstract}

We explore and expand the \textit{Soft Nearest Neighbor Loss} to measure the \textit{entanglement} of class manifolds in representation space: i.e., how close pairs of points from the same class are relative to pairs of points from different classes. We demonstrate several use cases of the loss. As an analytical tool, it provides insights into the evolution of class similarity structures during learning. Surprisingly, we find that \textit{maximizing} the entanglement of representations of different classes in the hidden layers is beneficial for discrimination in the final layer, 
possibly because it encourages
representations to identify class-independent similarity structures. 
Maximizing the soft nearest neighbor loss  in the hidden layers leads not only to
improved generalization but also to better-calibrated estimates of uncertainty on outlier data.
Data that is not from the training distribution can be recognized by observing that in the hidden layers, it has fewer than the normal number of neighbors from the predicted class.

\end{abstract}

\section{Introduction}

From SVM kernels to hidden layers in neural nets, the similarity structure  of representations 
plays a fundamental role in how well classifiers generalize from  training data. 
Representations are also instrumental in enabling well-calibrated confidence estimates for model
predictions.
This is particularly important when the model is likely to
be presented with outlier test data: {\it e.g.} to assist with
medical diagnostics when a patient has an unknown condition, 
or more generally when safety or security are at stake.

In this paper, we use the labels of the data points to illuminate the class similarity structure of the internal representations learned by discriminative training.
Our study of internal representations is structured around a loss function, 
the \textit{soft nearest neighbor loss}~\cite{salakhutdinov2007learning}, which we explore to measure the lack of separation of class manifolds in representation space---in other words,
the \textit{entanglement} of different classes. 
We expand upon the original loss by introducing a notion of temperature to
control the perplexity at which entanglement is measured.
We show several use cases of this loss including
as an analytical tool for the progress of discriminative and generative training. 
It can also be used to measure the entanglement of synthetic and real data in generative tasks.
 
We focus mainly on the effect of deliberately {\it maximizing} the entanglement of hidden representations in a classifier.  Surprisingly, we find that, unlike the penultimate layer, hidden layers that perform feature extraction benefit from being entangled. That is, they should \textit{not} be forced to disentangle 
data from different classes. In practice, we promote the entanglement of hidden layers
by adding our soft nearest neighbor loss as a bonus to the training objective.
We find that this bonus regularizes the model by encouraging representations that are already similar to become more similar if they have different labels.
The entangled representations form
class-independent clusters which capture other kinds of similarity that is helpful for eventual discrimination.  

In addition to this regularization effect, entangled representations
support better estimates of uncertainty on outlier data, such as 
adversarial examples or test inputs from a different distribution.
In our empirical study, we measure uncertainty with the Deep k-Nearest Neighbors (DkNN): the approach relies on a nearest neighbor search in the representation spaces of the model
to identify support in the training data for a given test input~\cite{papernot2018deep}.
Since entangled representations exhibit a similarity structure
that is less class-dependent, entangled models 
more coherently project outlier data
that does not lie on the training manifold. In particular, data that is not from the training distribution has fewer than the normal number of neighbors in the predicted class. As a consequence, 
uncertainty estimates provided by the DkNN are better 
calibrated on  entangled models.

The contributions of this paper are the following: 
\vspace*{-0.1in}
\begin{itemize}
    \item We explore and expand the soft nearest neighbor loss to characterize the class similarity structure in representation space (Section~\ref{sec:entanglement-loss}). Informally, the loss measures how entangled class manifolds are and can be used to track progress in both discriminative and generative tasks (Section~\ref{sec:entanglement-metric}).
    \item We show that {\it maximizing} representation entanglement by adding a bonus proportional to the soft nearest neighbor loss to the training objective serves as a regularizer (Section~\ref{sec:entangling}).
    \item We find that entangled representations deal better with outlier data far from the training manifold, thus supporting better confidence estimates on adversarial examples or different test distributions (Section~\ref{sec:entanglement-adversarial}).
\end{itemize}

\section{Soft Nearest Neighbor Loss}
\label{sec:entanglement-loss}

In the context of our work, the \textit{entanglement} of class manifolds characterizes how close pairs of representations from the same class are, relative to pairs of representations from different classes. If we have very low entanglement, then every representation is closer to representations in the same class than it is to representations in different classes. In other words, if entanglement is low then a nearest neighbor classifier based on those representations would have high accuracy. 

The \textit{soft nearest neighbor loss}~\cite{salakhutdinov2007learning} measures entanglement over labeled data. The loss computation can be approximated over a batch of data. 
Intuitively, we can think about this metric by imagining we are going to sample a neighboring point $j$ for every point $i$ in a batch, à la~\cite{goldberger2005neighbourhood},\footnote{The set of nearest neighbors for a given training point is also at the core of  unsupervised techniques for nonlinear dimensionality reduction like locally-linear embeddings~\cite{roweis2000nonlinear}.} 
where the probability of sampling $j$ depends on the distance between points $i$ and $j$. The soft nearest neighbor loss is the negative log probability of sampling a neighboring point $j$ from the same class as $i$. Our definition introduces a new parameter, the temperature, to control the relative importance
given to the distances between pairs of
points.

\paragraph{Definition.} The \textit{soft nearest neighbor loss} at temperature $T$, for a batch of $b$ samples $(x,y)$, is:
\begin{equation}
  \label{eq:ent-loss}
    l_{sn}(x,y,T) = -\frac{1}{b} \sum_{i\in 1..b} \log \left( \frac{\sum\limits_{\substack{j\in1..b \\j \neq i\ \\ y_i = y_j}}e^{-\frac{||x_i-x_j||^2}{T}}}{\sum\limits_{\substack{k\in1..b \\ k \neq i}} e^{-\frac{||x_i-x_k||^2}{T}}}\right)
\end{equation}
where $x$ may be either the raw input vector or its representation in some hidden layer. 
 At low temperatures, the loss is dominated by the small distances and the actual distances between widely separated representations are almost irrelevant. 
We include TensorFlow code outlining the matrix operations needed to compute this loss efficiently with our submission.

\begin{figure}[t]
\minipage{0.24\textwidth}
  \includegraphics[width=\linewidth]{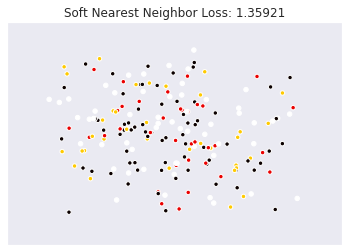}
\endminipage\hfill
\minipage{0.24\textwidth}
  \includegraphics[width=\linewidth]{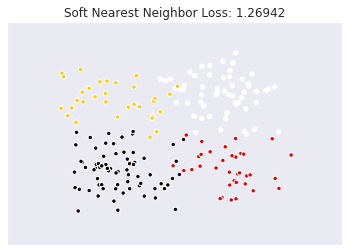}
\endminipage\hfill
\minipage{0.24\textwidth}%
  \includegraphics[width=\linewidth]{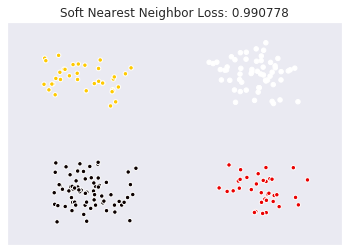}
\endminipage
\minipage{0.24\textwidth}%
  \includegraphics[width=\linewidth]{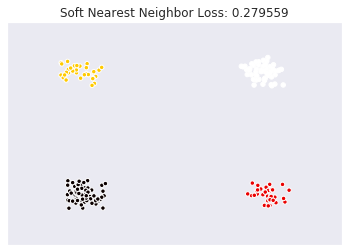}
\endminipage
\caption{A set of 200 2D points is sampled from a Gaussian and labeled randomly. Then, using gradient descent on the x and y coordinates of the points, the soft nearest neighbor loss is minimized to decrease entanglement. The 4 classes become more isolated.  While a direct comparison with other losses like cross-entropy is not possible for this experiment, we inspect and compare non-entangled and entangled representation spaces later in the paper.}
\label{fig:ent_vis_1}
\end{figure}

We plot different distributions annotated with their entanglement in Figure~\ref{fig:ent_vis_1}.
As we minimize the soft nearest neighbor loss to decrease entanglement, the  result is not necessarily each class collapsing to a {\it single} point. The loss is low when each point is closer to members of its own class than to other classes, but this can be achieved by having several widely separated pure cluster for each class.  This is illustrated in Figure~\ref{fig:ent_vis_2} (Appendix~\ref{ap:bimodal}) by introducing a second mode in each of the classes, which is preserved when entanglement is minimized by gradient descent on the soft nearest neighbor.

\begin{figure}[t]
\centering
\includegraphics[width=\linewidth]{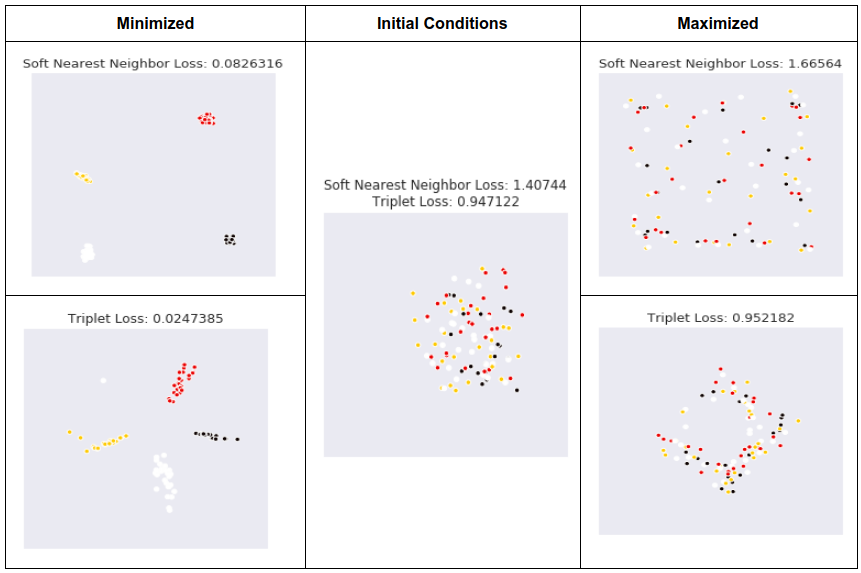}
\label{fig:triplet_vs_ent2}
\caption{Comparing the triplet and soft nearest neighbor losses. The middle plot shows the initial condition where each point is labeled by its color, the left plot shows the effect of minimizing either loss, and the right shows the effect of maximizing it.}
\end{figure}

Like the triplet loss~\citep{hoffer2015deep}, the soft nearest neighbor loss compares intra- to inter-class distances. 
However, a notable difference is that the triplet loss samples a single positive and negative point to estimate the separation of classes, whereas the soft nearest neighbor loss uses all positive and negative points in the batch. As visualized in Figure~\ref{fig:triplet_vs_ent2}: when maximizing the soft nearest neighbor loss, this results in representations that are  more spread out than the triplet loss. 
We show that this is a useful property of the soft nearest neighbor loss in Section~\ref{sec:entangling} and defer a more complete treatment of the triplet loss to Appendix~\ref{ap:triplet}. 

\paragraph{Temperature.}
By varying the temperature $T$, it is possible to alter the value of the loss function significantly. As outlined in Equation~\ref{eq:ent-loss}, temperature divides the squared distance between points before it is negatively exponentiated. Thus, when temperature is large, the distances between widely separated points can influence the soft nearest neighbor loss.  In the rest of this paper, we eliminate temperature as a hyperparameter by defining the entanglement loss as the minimum value over all temperatures:
\begin{equation}
    \label{eq:temp-ent-loss}
l^{\prime}_{sn}(x,y) = \arg\min_{T\in \mathbb{R}} l_{sn}(x,y,T)   
\end{equation}
 We approximate this quantity by initializing $T$ to a predefined value and, at every calculation of the loss, optimizing with gradient descent over $T$ to minimize the loss.\footnote{In practice, we found optimization to be more stable when we learn the inverse of the temperature.} 

\section{Measuring Entanglement during Learning}
\label{sec:entanglement-metric}

The soft nearest neighbor loss serves as an analytical tool to characterize the class similarity structure of representations throughout learning. In classifiers trained with cross-entropy, the soft nearest neighbor loss illuminates how models learn to compose entangled layers for  feature extraction with disentangled  layers for classification.  In generative models the loss shows how well they learn to entangle the synthetic data they generate with the real data from the distribution being modeled.

\begin{figure}[t]
  \centering
  \includegraphics[width=0.95\linewidth]{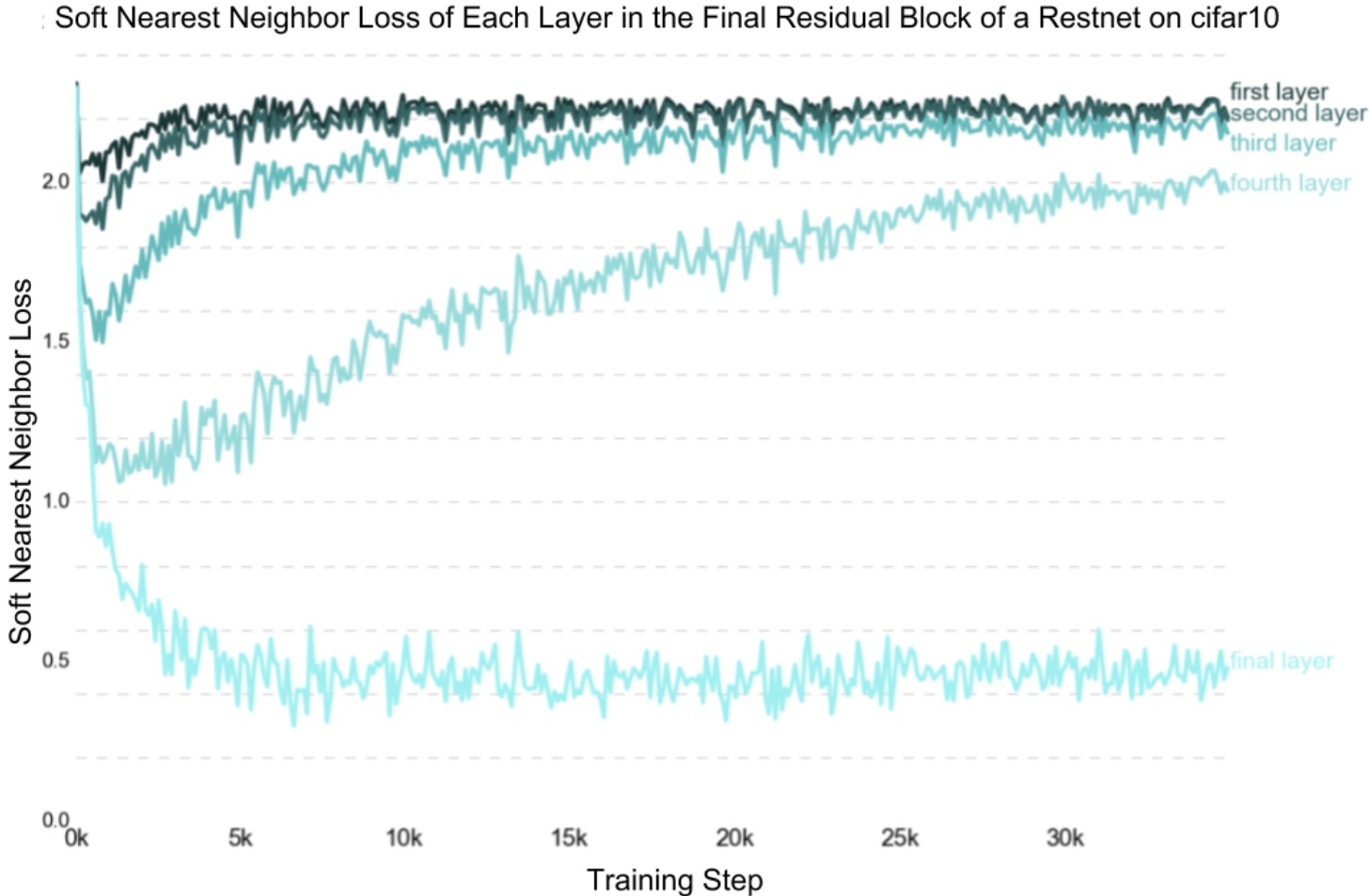}
  \caption{Entanglement of each layer within the last block of a ResNet on CIFAR-10, as measured with the soft nearest neighbor loss. Despite each layer initially disentangling data, as training progresses and features are co-opted as representations of sub features instead of classes, entanglement rises in all layers except for the final layer, which remains discriminative.}
  \label{fig:resnet}
\end{figure}

\subsection{Discriminative Models}
\label{ssec:discriminative-models}

With the soft nearest neighbor loss, we measure the entanglement of representations learned by each layer
 in the final block of a ResNet on CIFAR-10. 
In Figure~\ref{fig:resnet}, we distinguish two regimes. After an initial sharp decrease, the entanglement of lower layers of the block increases as training progresses. This suggests that the lower layers are discovering features shared by multiple classes.
By contrast, the entanglement of the block's output layer consistently decreases throughout training because the last hidden layer must allow linear separation of the logit for the correct class from all the other logits.

Qualitatively consistent conclusions can be drawn at the granularity of blocks (rather than layer), as demonstrated by an experiment found in Appendix~\ref{ap:discriminative}. Later in Section~\ref{sec:entangling}, we build on this perhaps counter-intuitive finding and propose maximizing a soft nearest neighbor loss to regularize gradient descent on the cross-entropy loss.

\subsection{Generative Models}

We now turn to 
generative models, and verify that they eventually entangle synthetic data with real data.
Then, we demonstrate how the soft nearest neighbor loss can act as an alternative to existing training objectives, in particular effectively replacing the discriminator used in GANs when semantics are captured by a distance in the input domain.

\paragraph{Entanglement in GANs. }
Synthetic data generated by GANs should be be highly entangled with real data because the generator is trained against a discriminator whose task is to discriminate between synthetic and real data~\cite{goodfellow2014generative}. 
Here, we are no longer calculating the (self) entanglement of a training batch, but rather calculating the entanglement between a batch of real data %
and a batch of synthetic data. %
This comes down to applying the soft nearest neighbor loss on a data batch containing equal splits of real and synthetic points, labeled as `real' or `synthetic'.

\begin{figure}[p]
  \centering
  \includegraphics[width=\linewidth]{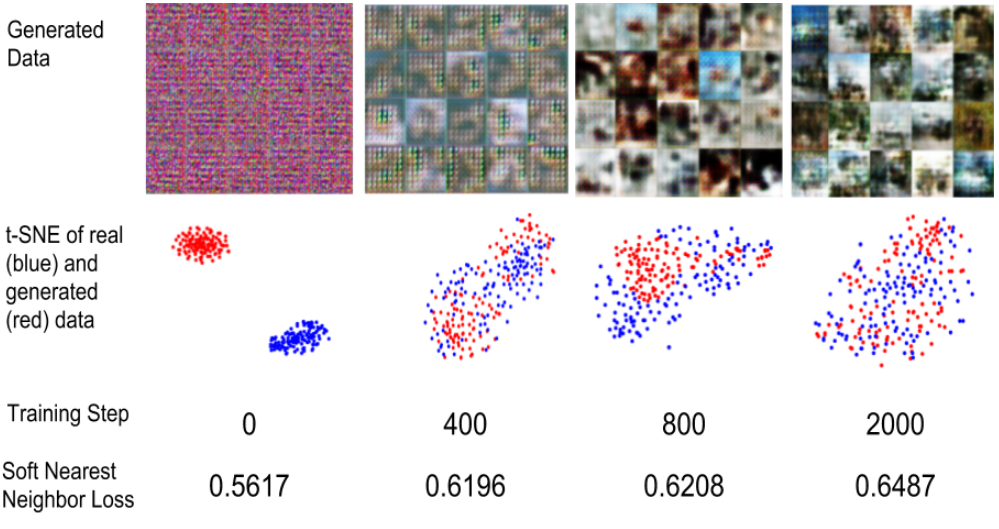}
  \caption{As training of vanilla GANs progresses, here on CIFAR10, the generator learns to entangle synthetic and training data, as confirmed by their increasing overlap in the t-SNE visualization as well as the larger soft nearest neighbor loss values. }
  \label{fig:gan-entanglement}
\end{figure}

In Figure~\ref{fig:gan-entanglement}, we report this measurement of entanglement  at different stages of training a GAN on CIFAR10. We also 
visualize real and synthetic data using t-SNE~\cite{maaten2008visualizing}. 
We observe that some modes of the input space are  ignored by the generator,
and conversely that some modes of the generated space are not representative of the true data distribution.
Note, however, how the real and synthetic data become less separable as training progresses, and how this is reflected in the entanglement score. This coherency between t-SNE and the soft nearest neighbor loss 
is to be expected given that both rely on similar calculations.

\begin{figure}[p]
  \centering
  \includegraphics[width=0.75\linewidth]{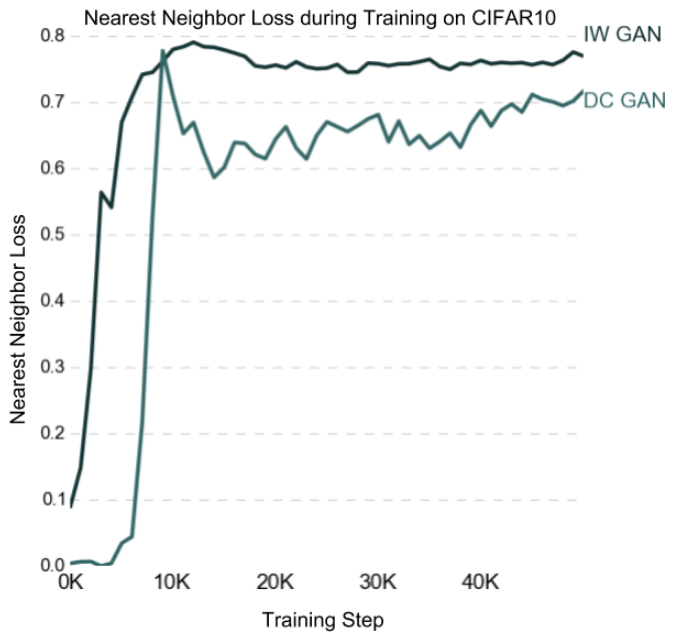}
  \caption{Entanglement of real and synthetic (generated) data throughout training, as measured with the soft nearest neighbor loss on two types of GAN architectures trained on CIFAR10.}
  \label{fig:cifar_gan}
\end{figure}

\begin{figure}[p]
  \centering
  \includegraphics[width=0.85\linewidth]{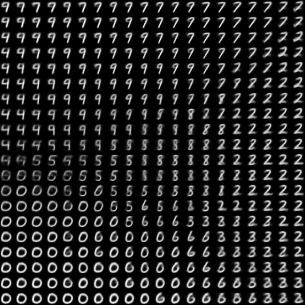}
  \caption{Images sampled from a generative model trained to maximize entanglement between synthetic and training data. The grid was created by extrapolating over 2 dimensions of the input noise.}
  \label{fig:ent_gan}
\end{figure}

Similarly to the aforementioned use of the soft nearest neighbor loss as a metric to evaluate class manifold separation during classifier training, we measure entanglement between the real  and  generated data throughout training. In the context of generative models, there is only one soft nearest neighbor loss evaluation per architecture, because entanglement is only defined in the input domain. In Figure~\ref{fig:ent_gan}, we see that two
variants of GANs exhibit different regimes of entanglement between synthetic and real CIFAR10 data as training progresses. We repeat the experiment on MNIST with qualitatively identical results in the Appendix~\ref{ap:gen}.

\paragraph{Soft nearest neighbor loss as an objective.}
\label{sec:gan-snnl-objective}

Given that GANs implicitly maximize entanglement, it is natural to ask whether the soft nearest neighbor loss can be used directly as a training objective for the generator.
To test this hypothesis, we replaced the discriminator (and its loss) with an inverse soft nearest neighbor loss in the GAN implementation used in our previous experiments on MNIST: i.e., the generator is now encouraged to maximize entanglement computed over a batch of real and synthetic data directly in pixel space.

On MNIST, this results in realistic and varied generated images (see Figure~\ref{fig:ent_gan}), which include all classes. Modes of the classes are captured by the generator, with for instance both the curly and straight ``2''. They are however noticeably smoother than data generated by traditional GANs. As a possible explanation, the generator maximizes the soft nearest neighbor loss evaluated on a batch when its output lies in between two training examples.
However, this strategy does not generalize to more complex datasets like CIFAR10, most likely because the Euclidean distance in pixel space used in the soft nearest neighbor loss does not adequately capture the underlying semantics of images. 

This limitation may most likely be overcome by measuring entanglement in a learned space, instead of pixel space. A potential preliminary instantiation of this intuition is explored in Appendix~\ref{ap:gan-sn}: we replace the cross-entropy loss that a normal discriminator minimizes with the soft nearest neighbor loss applied to a learned space. In this way, the discriminator learns a projection of the real and synthetic data that separates one from the other. 

Our proof-of-concept from Appendix~\ref{ap:gan-sn} demonstrates that this strategy succeeds on MNIST. 
This may also overcome the previously mentioned limitations for CIFAR10 image generation. 
However, our focus being classification, we leave a comprehensive investigation of the interplay between entanglement and generative modeling as future work.

\section{Entangling Representation Spaces}
\label{sec:entangling}

Apart from its characterization of similarity  in representation spaces, we found that the soft nearest neighbor may also serve as a training objective for generative models. At first, it appears that for discriminative models, one should encourage lower entanglement of internal representations by minimizing the soft nearest neighbor loss. Indeed, this would translate to
larger margins between
different classes~\cite{elsayed2018large}. 

However, we show here that \textit{maximizing} entanglement---in addition to minimizing cross-entropy---regularizes learning.
Specifically, training a network to minimize cross-entropy and maximize soft nearest neighbor loss reduces overfitting and achieves marginally better test performance. In Section~\ref{sec:entanglement-adversarial}, we will furthermore show that it promotes a class similarity structure in the hidden layers that better separates in-distribution from out-of-distribution data.

\subsection{Intuition behind Maximizing Entanglement}

Clustering data based on its labels is a natural avenue
for learning representations that discriminate: once a 
test point is assigned to a cluster of training points, 
its label can be inferred. This is referred to as the cluster assumption in the semi-supervised learning literature~\cite{chapelle2009semi}. 
However if test data is not represented in one of these class-homogeneous clusters, the behaviour of the network and the subsequent predicted label may be inconsistent. We argue that projecting all points in a class to a homogeneous clusters can be harmful to generalization and robustness.

Instead, we propose
regularizing the model by maximizing entanglement (through the soft nearest neighbor loss) to
develop 
class-independent  
similarity structures. This not-only promotes
spread-out intraclass representations, but also turns out to be good for recognizing data that is not
from the training distribution by observing that in the hidden layers,
such data has fewer than the normal number of neighbors from the predicted class.

Concretely, we minimize an objective that balances a cross-entropy term on logits and a soft nearest neighbor term on each hidden representation with a hyper-parameter $\alpha<0$, we represent the network as a series of transformations $f^k$, where $f^1$ is the first layer and $f^k$ is the logit layer. 
\begin{equation}
    \label{eq:total-loss}
    l(f, x, y) = -\sum_j y_j \log f^k(x_j) + \alpha \cdot \sum_{i\in k-1} l^{\prime}_{sn}(f^i(x), y)
\end{equation}

This may seem counter-intuitive but we note that many regularizers take on the form of two seemingly mutually exclusive objectives. For example label smoothing~\citep{pereyra2017regularizing} can be thought of trying to train a network to make accurate and confident predictions, but not overly confident. Similarly, dropout prompts individual neurons to operate independently from other---randomly deactivated---neurons, while still learning features that can be meaningfully combined \citep{srivastava2014dropout}. 
Here, 
our training objective simultaneously minimizes cross-entropy and maximizes the soft nearest neighbor loss. In other words, the model is constrained to learn representations whose similarity structure facilitates classification (separability) but also entanglement of representations from different
classes (inseparability).

\subsection{Soft Nearest Neighbor Loss as a Regularizer}

We first measure the generalization of models
that maximize the soft nearest neighbor loss in 
addition to minimizing
cross-entropy.
We trained a convolutional network\footnote{The architecture we used was made up of two convolutional layers followed by three fully connected layers and a final softmax layer. The network was trained with Adam at a learning rate of 1e-4, a batch size of 256 for 14000 steps.} on MNIST, Fashion-MNIST and SVHN, as well as a ResNet\footnote{The ResNet v2 with 15 layers was trained for 106 epochs with a exponential decreasing learning rate starting at 0.4.} on CIFAR10.
Two variants of each model were trained with a different objective: (1) a \textit{baseline} with cross-entropy only and (2) an \textit{entangled} variant balancing both cross-entropy and the soft nearest neighbor loss as per Equation~\ref{eq:total-loss}. As reported in Table~\ref{tbl:ent-regularizer}, all entangled models outperformed their non-entangled counterparts to some extent.

While we note that baseline accuracies we report are below the current state-of-the-art for the corresponding datasets, this is an intentional experimental design choice we made. Indeed, we wanted to isolate the behavior of our soft nearest neighbor loss from other factors (e.g., dropout or other regularizers) that may impact representation spaces.

\begin{table}[p]
\begin{center}
\begin{tabular}{|l|l|l|l|}
\hline
CNN Model                                                                                   & Test Accuracy & Entangled       & Baseline \\ \hline
\multirow{2}{*}{\begin{tabular}[c]{@{}l@{}}MNIST\end{tabular}}           & Best          & \textbf{99.23\%} & 98.83\%   \\ \cline{2-4} 
                                                                                   & Average       & \textbf{99.16\%} & 98.82\%   \\ \hline
\multirow{2}{*}{\begin{tabular}[c]{@{}l@{}}Fashion-\\ MNIST\end{tabular}} & Best          & \textbf{91.48\%} & 90.42\%   \\ \cline{2-4} 
                                                                                   & Average       & \textbf{91.06\%} & 90.25\%  \\ \hline
\multirow{2}{*}{\begin{tabular}[c]{@{}l@{}}SVHN\end{tabular}}            & Best          & \textbf{88.81\%} & 87.63\%   \\ \cline{2-4} 
                                                                                  & Average       & \textbf{89.90\%} & 89.71\%   \\ \hline
\hline
ResNet Model                                                                                & Test Accuracy & Entangled       & Baseline \\ \hline
\multirow{2}{*}{\begin{tabular}[c]{@{}l@{}}CIFAR10\end{tabular}}     & Best          & \textbf{91.220\%} & 90.780\%   \\ \cline{2-4} 
                                                                                   & Average       & \textbf{89.900\%} & 89.713\%   \\ \hline
\end{tabular}
\end{center}
\caption{Using a composite loss, which minimizes cross entropy loss and maximizes entanglement through the soft nearest neighbor loss, marginally increases test performance on all datasets studied. A CNN was used for MNIST, FashionMNIST and SVHN. ResNet was used for CIFAR10. Values are averaged over 4 runs for the CNN and 100 runs for the ResNet. No additional regularizers  were used and thus we achieve less than state-of-the-art performance, but are able to study the Soft Nearest Neighbor loss in isolation.}
\label{tbl:ent-regularizer} 
\end{table}

\begin{figure}[p]
\centering
\includegraphics[width=0.85\linewidth]{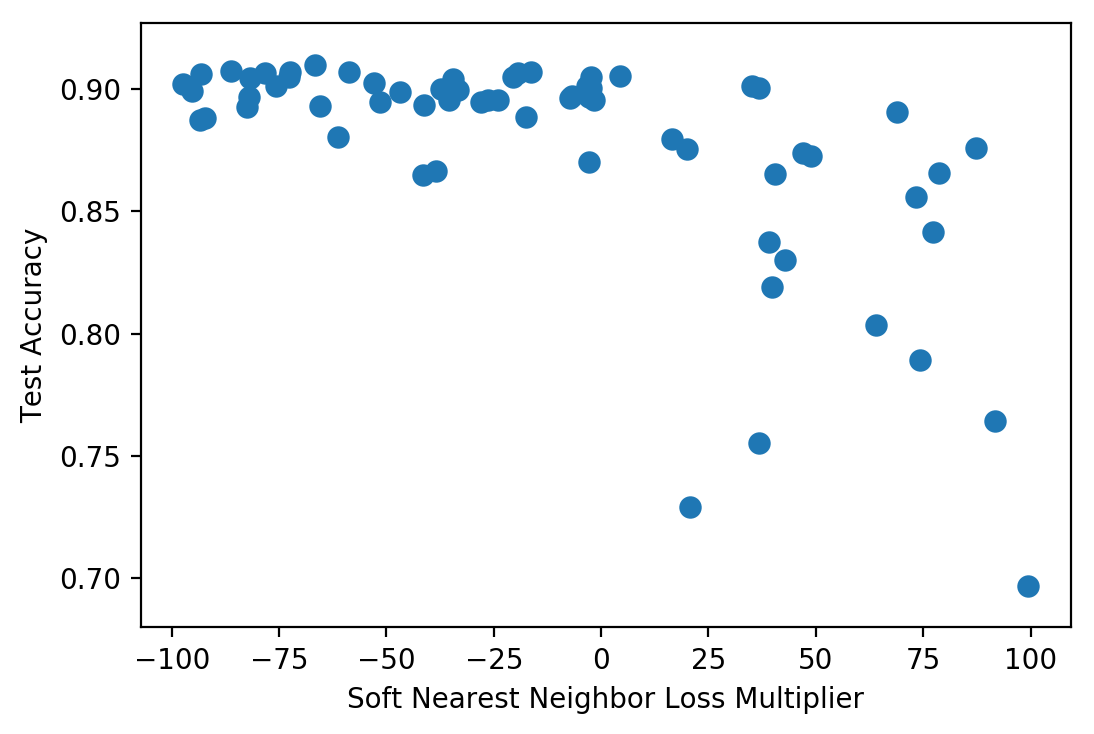}
\caption{Test accuracy as a function of the soft nearest neighbor hyper-parameter $\alpha$ for 64 training runs of a ResNet v2 on CIFAR10. These runs are selected by a strategy to tune the learning rate, entanglement hyper-parameter $\alpha$, and initial temperature $T$.}
\label{fig:VizierRunsAccVsEntMult}
\end{figure}

\begin{figure}[p]
\centering
\includegraphics[width=\linewidth]{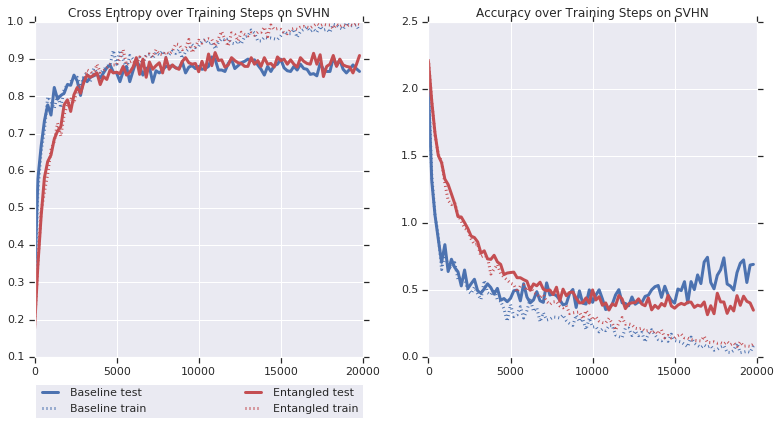}
\caption{Accuracy and cross-entropy for baseline (blue) and entangled (red) models as a function of the number of training steps. In addition to increased test accuracy (left), the smaller gap between cross-entropy on training and test data (right) for entangled models illustrates how they begin to overfit much later and to a much lesser degree than  non-entangled counterparts. Curves are averaged over two runs for both models.}
\label{fig:SVHN_v_steps_entanglement}
\end{figure}

To validate that maximizing entanglement is beneficial for generalization, we fine-tuned the hyperparameter $\alpha$ that balances the cross-entropy and soft nearest neighbor terms in our objective. The search was conducted on our CIFAR10 model using a strategy based on Batched Gaussian Process Bandits~\cite{desautels2014parallelizing}. Because both positive and negative values of $\alpha$ were considered, this search explored respectively both minimization and maximization of representation entanglement. As illustrated by Figure~\ref{fig:VizierRunsAccVsEntMult}, the search independently confirmed that maximizing entanglement led to better test performance as it eventually converged to large negative values of $\alpha$.

To explain the increased test performance of entangled models, we hypothesized that the entanglement term added to our training objective serves as a regularizer. To verify this, we measured the cross-entropy loss on  training and test 
data while training the non-entangled and entangled variants of our models for a large number of steps. This allowed for overfitting.
We draw the corresponding learning curves for SVHN in Figure~\ref{fig:SVHN_v_steps_entanglement} and observe that the entangled model not only overfits at a later stage in training (about 5,000 steps later), it also overfits to a much lesser degree.

\section{Entangled Models in Adversarial Settings}
\label{sec:entanglement-adversarial}

Given the improved---more class-independent---similarity structure of entangled representations obtained through maximizing the soft nearest neighbor loss,  we hypothesize that entangled models also offer better estimates of their uncertainty. Here, we do \textit{not} claim robustness to adversarial examples but rather show that entangled representations help distinguish outliers from real data. 
We validate this by considering two types of out-of-distribution test data: first, maliciously-crafted adversarial examples, and second, real inputs from a different test distribution. We find that hidden layers of entangled models consistently represent outlier data far away from the expected distribution's manifold.

It is natural to ask if  reduced class margins make entangled representations more vulnerable to adversarial perturbations. This is not necessarily the case. In fact, we show in  Appendix~\ref{ap:adv} that models with state-of-the-art robustness on MNIST have higher  entanglement than  non-robust counterparts.
Furthermore, recent work has found that when models concentrate data, they are more vulnerable to adversarial examples~\cite{mahloujifar2018curse}, whereas entangled models encourage intraclass clusters to spread out.

\vspace*{-0.05in}

\paragraph{Attack techniques.}
Our study considers both white-box and black-box threat models. 
Given access to gradients in the white-box setting, various heuristics and optimization algorithms allow the adversary to
create adversarial examples
~\citep{biggio2013evasion,szegedy2013intriguing}. Here, we use both single-step and iterative attacks: the Fast Gradient Sign Method~\cite{goodfellow2014explaining} and Basic Iterative Method~\cite{kurakin2016adversarial}. When
gradients are unavailable, as is the case
for black-box interactions
(i.e., the
adversary only has access to the label predicted), a common strategy is to first find adversarial examples
on a substitute model and then transfer them to the victim model~\citep{szegedy2013intriguing,papernot2017practical}.
Adversarial perturbations are said to be \textit{universal} if they change a model's prediction into a chosen class
once added to \textit{any} input \citep{goodfellow2014explaining,moosavi2017universal}.

\vspace*{-0.05in}

\paragraph{Uncertainty estimation.} Estimating the epistemic uncertainty that stems from the finite nature of datasets analyzed by models during learning remains an open problem. In our work, we apply a recent proposal called the Deep k-Nearest Neighbors~\cite{papernot2018deep} that computes the credibility of each test-time prediction; a metric that reflects how well the training data supports this prediction. The approach consists in running a k-nearest neighbors search in the representation space learned by each hidden layer so as to extract the k training points whose representation is closest to the predicted representation of the test point considered. If the labels of these nearest training points largely agree with the test label being predicted, the prediction is assigned high credibility. Otherwise, it is assigned a low credibility score, which implies it should not be relied upon. A holdout dataset is used to calibrate the expected level of agreement between the training and test data.

\subsection{Entangled Representations support more Calibrated DkNN Estimates of Uncertainty}
\label{ssec:sn-dknn}

In the original proposal,  the DkNN is applied to vanilla neural networks without modifying the way they are trained. Intuitively, training with the soft nearest neighbor loss should impact the credibility predicted by the DkNN because it modifies the class similarity structure of hidden representations that are core to the analysis performed by the DkNN.

\begin{figure}[t]
\centering
\includegraphics[width=\linewidth]{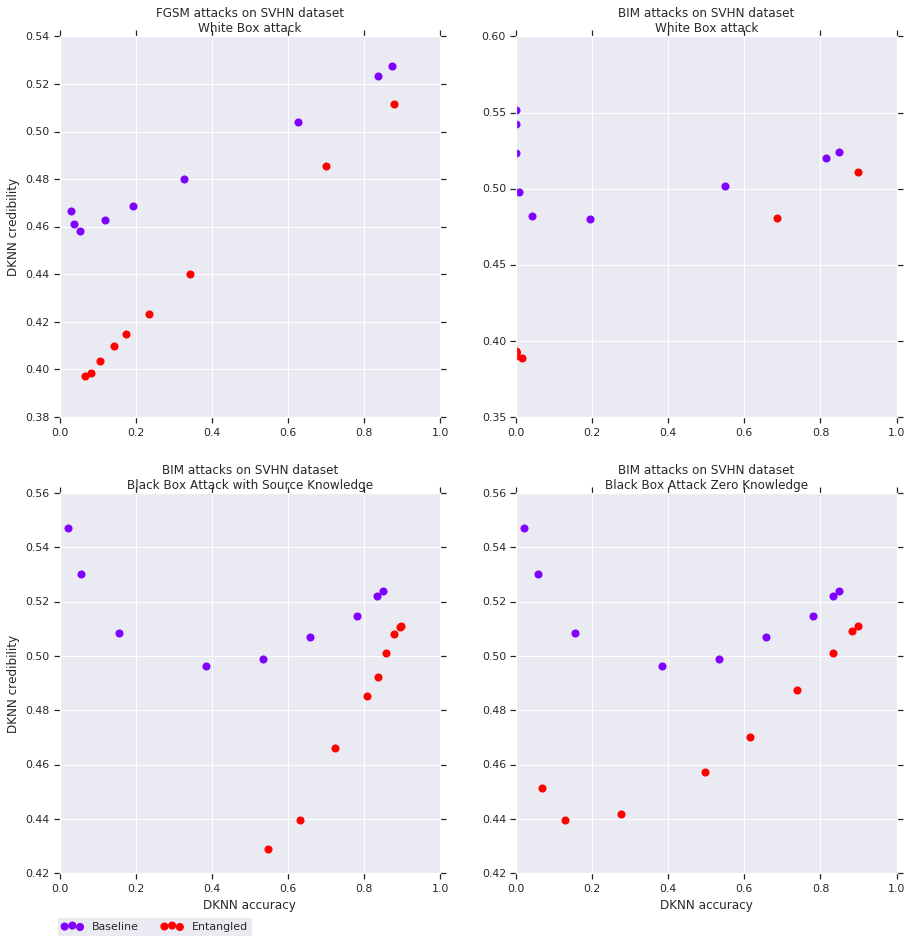}
\caption{DkNN credibility (i.e., uncertainty estimate) as a function of prediction accuracy on SVHN (averaged over two runs). Each point corresponds to adversarial examples generated with $\varepsilon \in [0.01, 0.5]$. Iterative attacks use a fixed number of steps (1000) and a fixed $\alpha = 0.01$ for all $\varepsilon$. Plots are shown for white-box FGSM attack (top left), white-box BIM attack (top right), black-box attacks with source knowledge (bottom left), black-box attacks with zero knowledge (bottom right). Source knowledge implies the adversary is aware of the defense and transfers adversarial examples from a model trained with the same loss, whereas zero knowledge adversaries always transfer from a model trained with cross-entropy. This allows us to rule out most common forms of gradient masking. In all cases, entangled models yield credibility estimates that are more correlated with accuracy, and the two bottom graphs show that they suffer less from transferability.}
\label{fig:dknn_svhn}
\end{figure}

Using MNIST, Fashion-MNIST and SVHN, we compare two models : one trained with cross-entropy only and one with the composite loss from Equation~\ref{eq:total-loss} that includes a cross-entropy term and soft nearest neighbor term. 
We compare how the two models' credibility estimates correlate with their predictive accuracy. Ideally, the relationship between the two should be the identity; if a DkNN system was perfectly calibrated then inputs that were correctly classified would have 100\%  credibility while inputs that were incorrectly classified would have 0\% credibility.

We tested each model on FGSM and BIM adversarial examples assuming white-box access to the model, with progressively larger perturbations ($\varepsilon$ gradient step). We also considered adversarial examples crafted with the BIM attack but transferred from a different model. This black-box attack enables us to test for gradient masking. In Figure~\ref{fig:dknn_svhn}, we then plotted the average DkNN credibility (low credibility corresponds to higher uncertainty) with respect to the classification accuracy. Each point corresponds to a different  perturbation magnitude. While the credibility is not perfectly linear with respect to the accuracy for either the standard or entangled model, the correlation between credibility and accuracy is consistently higher for entangled models in both the white-box and black-box settings.

To explain this, we t-SNE representations in Appendix~\ref{ap:intuition} and find that entangled models better separate adversarial data from real data in activation space. This in turn implies that adversarial data can be recognized as not being part of the distribution by observing that it has fewer than the normal number of neighbors from the predicted class.

\subsection{Transferability and Representation Entanglement}
\label{ssec:Transferability}

Transferability---the fact that adversarial examples for one model are also often misclassified by a different model---was empirically found to apply to a wide range of model pairs, despite these models being trained with different ML techniques (e.g., decision trees and neural nets) or subsets of data. Several hypotheses were put forward to explain why this property holds in practice, including gradient alignment.

\begin{figure}[p]
\centering
\includegraphics[width=0.75\linewidth]{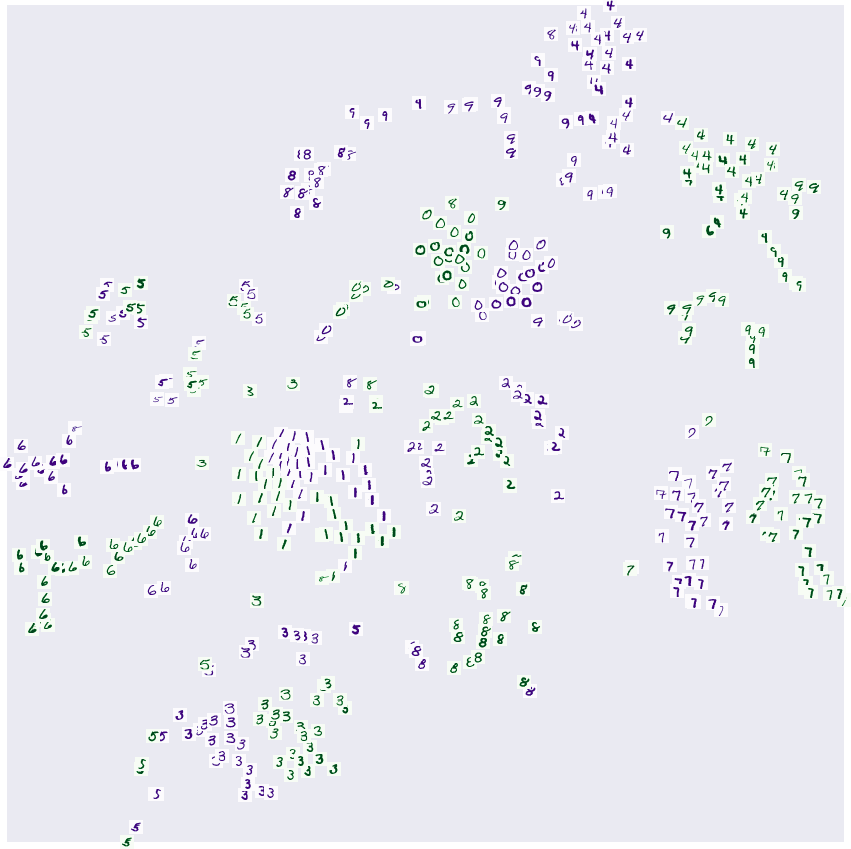}
\caption{t-SNE visualization of  gradients computed by a FGSM attack targeting class 1 on two vanilla models, one in green the other in blue.}
 \label{fig:tsne_gradients}
\end{figure}

This is visualized in Figure~\ref{fig:tsne_gradients}, which plots gradients followed by a targeted FGSM attack in two dimensions using t-SNE. The plot stacks the visualizations for two different models.  One can see that coherent clusters exist across the two individual models. This means that gradients that are adversarial to one model are likely to be aligned with gradients that are adversarial to a second model.

\begin{figure}[p]
\centering
\includegraphics[width=0.75\linewidth]{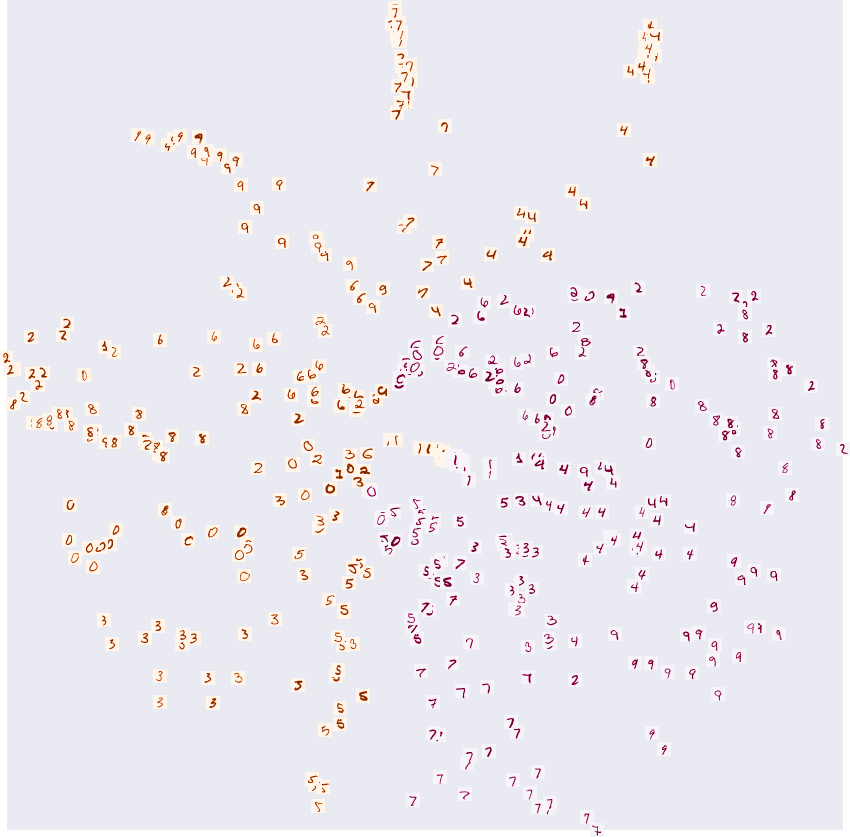}
\caption{t-SNE visualization of  gradients computed by a FGSM attack targeting class 1 on two entangled models, one in red the other in orange.}
 \label{fig:tsne_gradients_entangled}
\end{figure}

However, this gradient alignment does not hold in entangled models. When we repeat the same experiment with a standard cross-entropy model and an entangled model, or two entangled models, the clusters are no longer coherent across pairs of models---as illustrated in Figure~\ref{fig:tsne_gradients_entangled}. This suggests that while adversarial examples can still be found in the white-box setting by following the gradients of a specific entangled model, it is harder to find perturbations that are universal (i.e., apply to any test input) or transferable (i.e., apply across different entangled models).

\subsection{Out-of-Distribution Test Inputs}
\label{ssec:outlier-data}

\begin{figure}[p]
\centering
\includegraphics[width=0.49\linewidth]{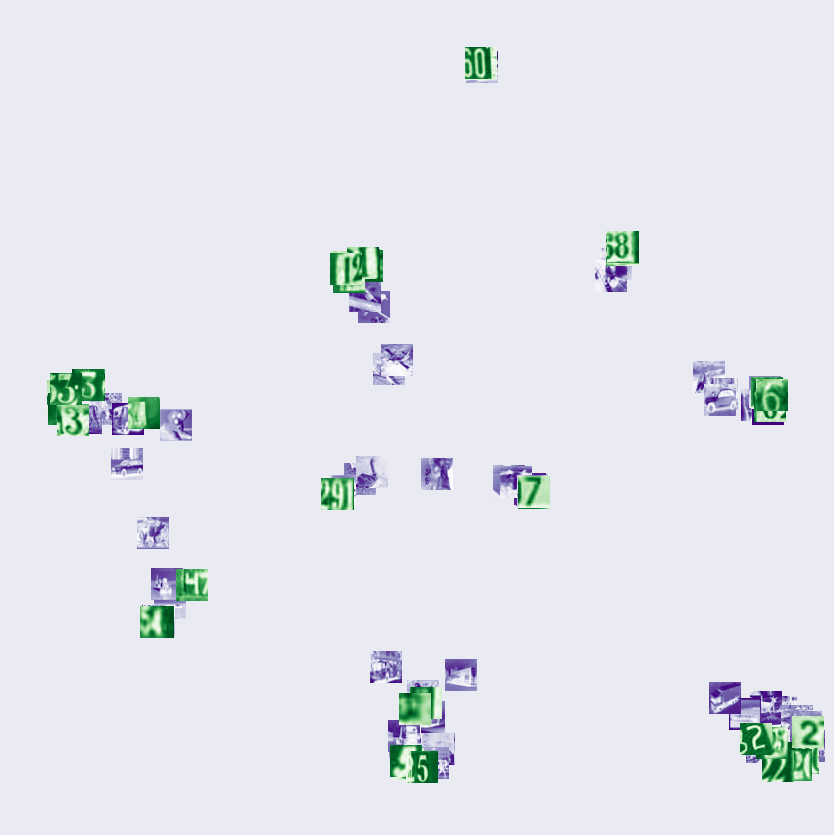}
\includegraphics[width=0.49\linewidth]{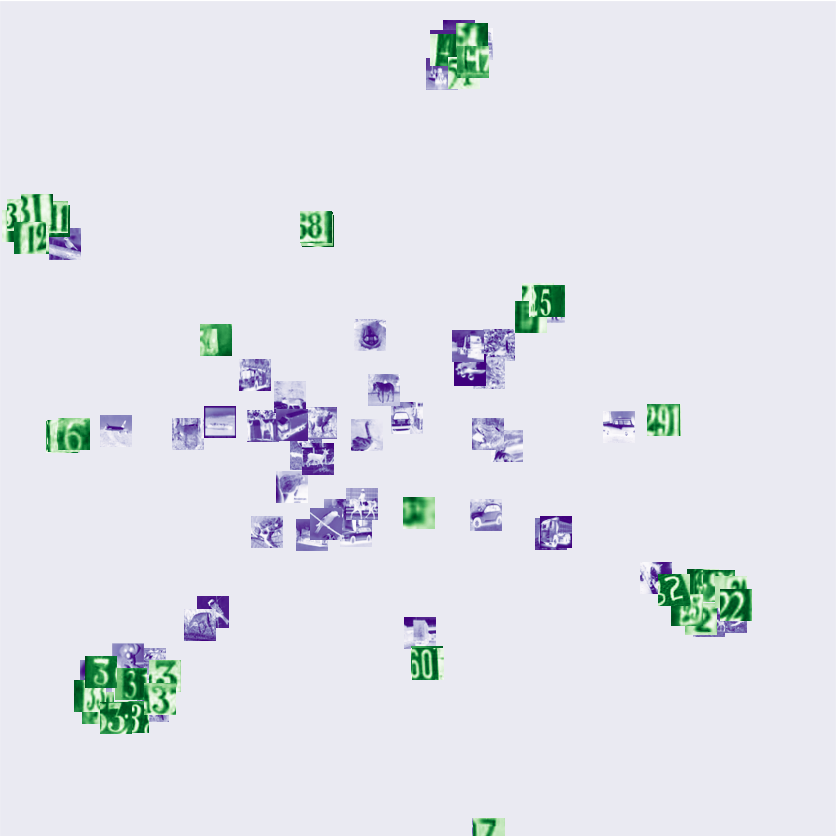}
\caption{t-SNE visualization of logits for in-distribution (SVHN--- green) and out-of-distribution (cifar10---blue) test data learned by a baseline (left) and entangled (right) model. We can see that the out of distribution data is easier to separate from the true data for the entangled model than it is for the baseline model.}
\label{fig:svhn_cifar10}
\end{figure}

Unlike techniques like adversarial training~\cite{szegedy2013intriguing}, training with the soft nearest neighbor loss relies only on the original training data and makes no assumptions about a particular algorithm used to generate the out-of-distribution examples. 
Hence, having shown that training a network to maximize entanglement  leads to representations that better separate adversarial data from real data, we expect this behaviour to be consistent across any data sampled from something other than the expected test distribution. This includes inputs from a different test distribution.

To test this we can train a network on SVHN and see what its behavior is like on CIFAR10: test examples from  CIFAR10  should be represented very differently  from the SVHN test examples. This is indeed what we observe in Figure~\ref{fig:svhn_cifar10}, which uses t-SNE to visualize how the logits represent SVHN and CIFAR10 test inputs when a model is trained with cross-entropy only or with the soft nearest neighbor loss to maximize entanglement. The vanilla model makes confident predictions in the SVHN classes for the CIFAR10 inputs (because they are represented close to one another), whereas the entangled model separates all of the CIFAR10 data in a distinct cluster and preserves the SVHN clusters. A similar experiment on a MNIST model using notMNIST as out-of-distribution test inputs is found in  Appendix~\ref{ap:outlier}.

\section{Conclusions}

We expanded on and explored novel use cases of the soft nearest neighbor loss. It can serve as a tool to characterize the class similarity structure of representations, allowing us to measure learning progression of discriminative models. The loss also captures how generative models entangle synthetic and real data, and can thus serve as a generative loss itself. Furthermore, by adding the loss as a bonus to a classifier's training objective, we are able to boost test performance and generalization. Because entangled representations are encouraged to spread out data further in activation space (see Figure~\ref{fig:triplet_vs_ent2}), they represent outlier data more consistently apart from real data (see Figure~\ref{fig:adv-data-projection}). This in turn means outlier data is easily rejected by observing that it is supported by fewer neighbors from the predicted class, as captured by our improved uncertainty estimates. 

 \subsubsection*{Acknowledgments}
The authors would like to thank Martin Abadi, Samy Bengio, Nicholas Carlini, Yann Dauphin, Ulfar Erlingsson, Danijar Hadner, Ilya Mironov, Sara Sabour, Kunal Talkwar and Nithum Thain for insightful comments on this project.

\bibliography{references}
\bibliographystyle{icml2018}

\newpage

\begin{appendices}

\section{Soft Nearest Neighbor Loss on Toy distribution}
\label{ap:bimodal}

This Figure complements Figure~\ref{fig:ent_vis_1}. It adds a second mode to each class of the distribution, showing that minimizing entanglement through the soft nearest neighbor loss preserves the two modes in each class.

\begin{figure}[h!]
\minipage{0.24\textwidth}
  \includegraphics[width=\linewidth]{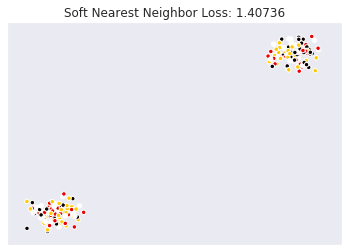}
\endminipage\hfill
\minipage{0.24\textwidth}
  \includegraphics[width=\linewidth]{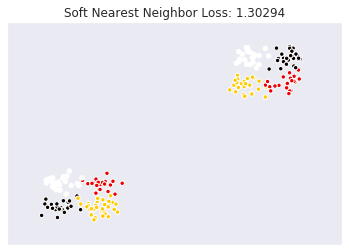}
\endminipage\hfill
\minipage{0.24\textwidth}%
  \includegraphics[width=\linewidth]{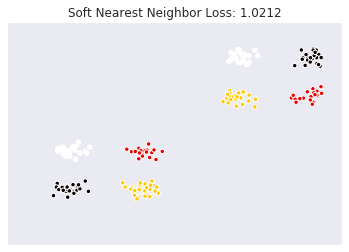}
\endminipage
\minipage{0.24\textwidth}%
  \includegraphics[width=\linewidth]{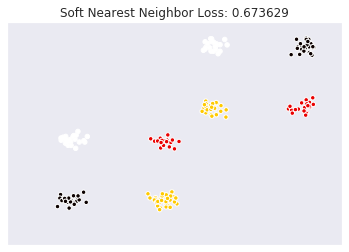}
\endminipage
\caption{Data is generated for each class by sampling from two Gaussians. As entanglement is minimized using gradient descent on the $(x, y)$ coordinates of the points, each class does not collapse into a single point; instead, both initial modes are preserved.}\label{fig:ent_vis_2}
\end{figure}

\section{Comparing the Soft Nearest Neighbor Loss with the Triplet Loss}
\label{ap:triplet}

The soft nearest neighbor loss is similar to the triplet loss~\citep{hoffer2015deep},
in that both measure the relative distance between points from the same
class and points from different classes. 
The triplet loss is calculated by taking the maximum of 0 and the difference between (a) the distance between an anchor point and a positive point (in the same class) and (b) the distance between the anchor point and a negative point (in a different class) for every anchor point in a batch. Equation~\ref{eq:triplet_loss} presents the triplet loss where $x_i^a$ denotes the anchor point, $x_i^a$ a positive sample, $x_i^n$ a negative one and $\alpha$ the margin term: 
\begin{equation}
    \label{eq:triplet_loss}
    L = \sum_i^N \big(||f(x_i^a) - f(x_i^p)||_2^2 - ||f(x_i^a) - f(x_i^n)||_2^2 + \alpha \big)
\end{equation}

Minimizing the triplet loss should have a similar effect on learned representations as minimizing entanglement (by minimizing the soft nearest neighbor loss) as both are imposing constraints on the relative distance between points within a class and points in different classes.
However, a notable difference is that the triplet loss is calculated by sampling positive and negative points to estimate the separation of classes whereas the soft nearest neighbor loss uses all of the points in a batch to measure the separation.

In Figure~\ref{fig:triplet_vs_ent2}, we compare minimizing and maximizing these two similar losses by visualizing the results of minimizing and maximizing a random set of 2 dimensional points labeled in four classes. We see that both losses have similar effects when the loss is minimized: the classes are separated by a larger margin. However, when the loss is maximized, the end results are not identical: the triplet loss chooses a representation that densely projects the data around a circle whereas the soft nearest neighbor loss spreads out data throughout the representation space.

\begin{figure}[t]
\centering
\includegraphics[width=\linewidth]{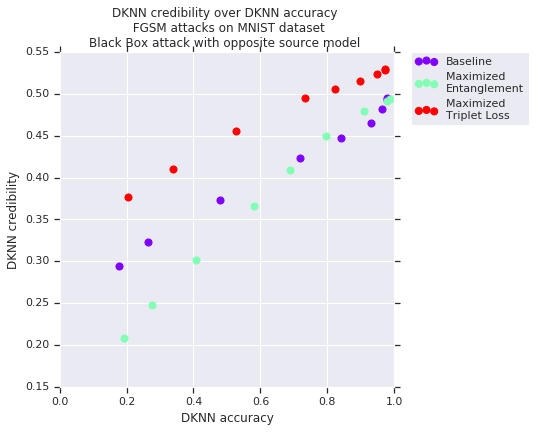}
\caption{DKNN credibility over accuracy for white box FGSM attacks with varying epsilons, plotted for MNIST. Maximizing the triplet loss seems to have the opposite effect of maximizing the soft nearest neighbor loss.}
\label{fig:dknncred_MNIST_triplet}
\end{figure}

We provide an additional point of comparison: the impact of both losses on the calibration of DkNN credibility estimates. We train MNIST models  with cross-entropy and a regularizer maximizing either the triplet loss or the soft nearest neighbor loss at each layer, as done in Section~\ref{ssec:sn-dknn}. We report the accuracy of DkNN predictions with respect to their credibility in Figure~\ref{fig:dknncred_MNIST_triplet}.  We did not find improved DKNN calibration for networks trained with the triplet loss term---unlike models maximizing entanglement through the soft nearest neighbor term.

\section{Additional Entanglement Measurements}
\label{ap:discriminative}

We report here entanglement measurements made with the soft nearest neighbor loss on MNIST and CIFAR10 models. They complement results presented in Section~\ref{ssec:discriminative-models}, which demonstrated the use of the soft nearest neighbor loss as an analytical tool to follow the evolution of similarity structures during learning in models trained to minimize cross-entropy.

\begin{figure}
  \centering
  \includegraphics[width=\linewidth]{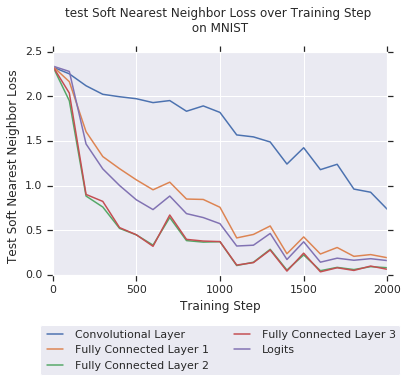}
  \caption{Soft nearest neighbor loss value per layer of a neural network on MNIST. The loss decreases during training despite the model being optimized to  minimize cross-entropy only.}
  \label{fig:MNIST_ent}
\end{figure}

\paragraph{MNIST.} We trained a neural network with one convolutional layer and three fully-connected layers on MNIST and measured the Soft Nearest Neighbor Loss of each training batch at each layer during training. Note in Figure~\ref{fig:MNIST_ent} how the loss value decreases throughout training, unlike results presented in Section~\ref{sec:entanglement-metric}. This is most likely because MNIST is easier to separate in the input domain than other datasets considered in our work. 

\begin{figure}[t]
  \centering
  \includegraphics[width=\linewidth]{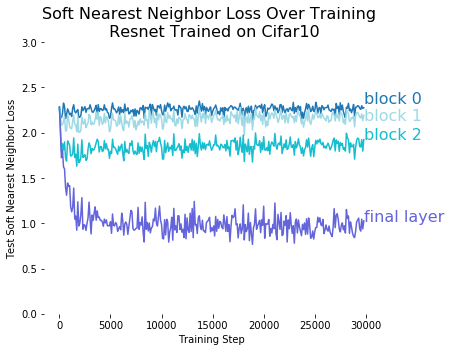}
  \caption{Entanglement loss averaged within each residual block of a ResNet on CIFAR-10. Entanglement remains high throughout learning for the lower blocks, as they extract features that help discriminate classes, and only decreases in the final block.}
  \label{fig:resnet-all}
\end{figure}

\paragraph{CIFAR10.} 
We repeat the experiment presented in Section~\ref{ssec:discriminative-models} but now looking at all  residual blocks instead of only the last one. In Figure~\ref{fig:resnet-all}, we report the average soft nearest neighbor loss of the  layers contained in each residual block,
across all of the training data throughout learning. Results are consistent with Section~\ref{ssec:discriminative-models}.
Entanglement is fairly constant or increases as training progresses in the first three blocks; suggesting  a large amount of feature co-adaptation across classes in the corresponding layers. 
Instead, the final block's entanglement  monotonically decreases as it extracts discriminative features to classify the input. When measuring Soft Nearest Neighbor Loss within a resnet with large hidden layers, we use cosine distance ($1 - cos(\pmb x, \pmb y)$) instead of euclidean distance to ensure stable calculations.

\section{Soft Nearest Neighbor Loss as an Analytical Tool for Generative Models}
\label{ap:gen}

\begin{figure}
  \centering
  \includegraphics[width=\linewidth]{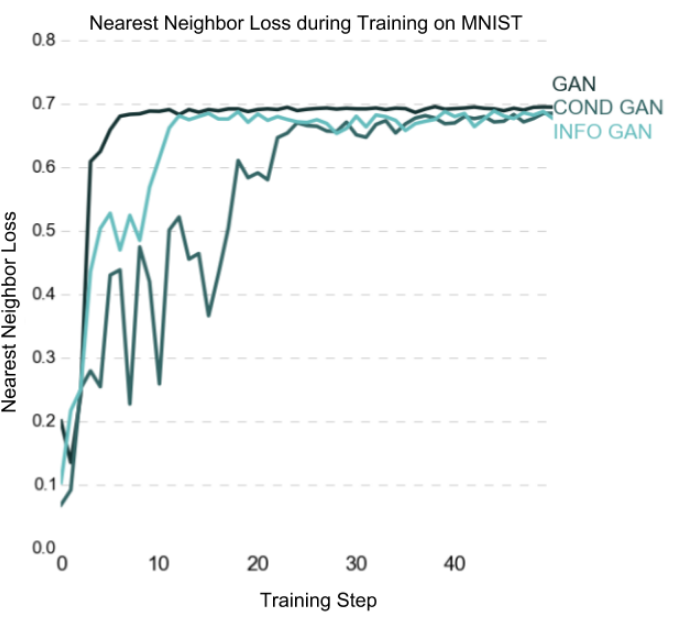}
  \caption{Entanglement of synthetic and real data, as measured by the soft nearest neighbor loss, on three different types of GAN architectures trained on MNIST.}
  \label{fig:mnist_gan}
\end{figure}

In Section~\ref{sec:entanglement-metric}, we showed how the soft nearest neighbor loss allows us to monitor the entanglement of synthetic data with real training data when learning a generative model on CIFAR10. Here, Figure~\ref{fig:mnist_gan} is the analog of Figure~\ref{fig:cifar_gan} for the MNIST dataset: it plots the entanglement  between synthetic and real data, as measured by the soft nearest neighbor loss, on three variants of GANs.  

\section{Does Entanglement conflict with Robustness?}
\label{ap:adv}

We reproduce the adversarial training procedure from~\citet{madry2017towards}, where adversarial examples are generated with projected gradient descent (that is with multiple gradient steps and random restarts).
The training objective \textit{only} minimizes cross-entropy over these adversarial examples. 
Once the model is trained, we measure the entanglement of its hidden layers using the soft nearest neighbor loss. The same architecture, also trained to minimize cross-entropy but on  \textit{non-adversarial} data, serves as a baseline to interpret these entanglement measurements. As reported in Table~\ref{tbl:entangle-madry-et-al}, we find that the adversarially trained model's convolutional layers are more entangled than the baseline model's, despite not being explicitly constrained to maximize entanglement during training. This further supports our hypothesis that increased entanglement of representation spaces is beneficial to the similarity structure  of internal representations and can support better (here, worst-case) generalization.

\begin{table}[t]
\begin{center}
\begin{tabular}{ c c c }
\textbf{Layer} & \textbf{Baseline model} & \textbf{PGD model} \\
Conv1 (after pool) & 1.39 & 2.21 \\
Conv2 (after pool) & 0.75 & 1.97 \\
Fully Connected layer & 1.75 & 0.46 \\
Logits & 0.13 & 0.21
\end{tabular}
\end{center}
\caption{Entanglement loss measured on models trained to minimize cross-entropy on the original training data (baseline model) or adversarial examples (PGD model). Measurements were made at temperature $T=100$ on a batch of 128 MNIST test points.}
\label{tbl:entangle-madry-et-al} 
\end{table}

\section{DkNN Uncertainty Calibration}

\begin{figure*}[t]
  \centering
  \includegraphics[width=\linewidth]{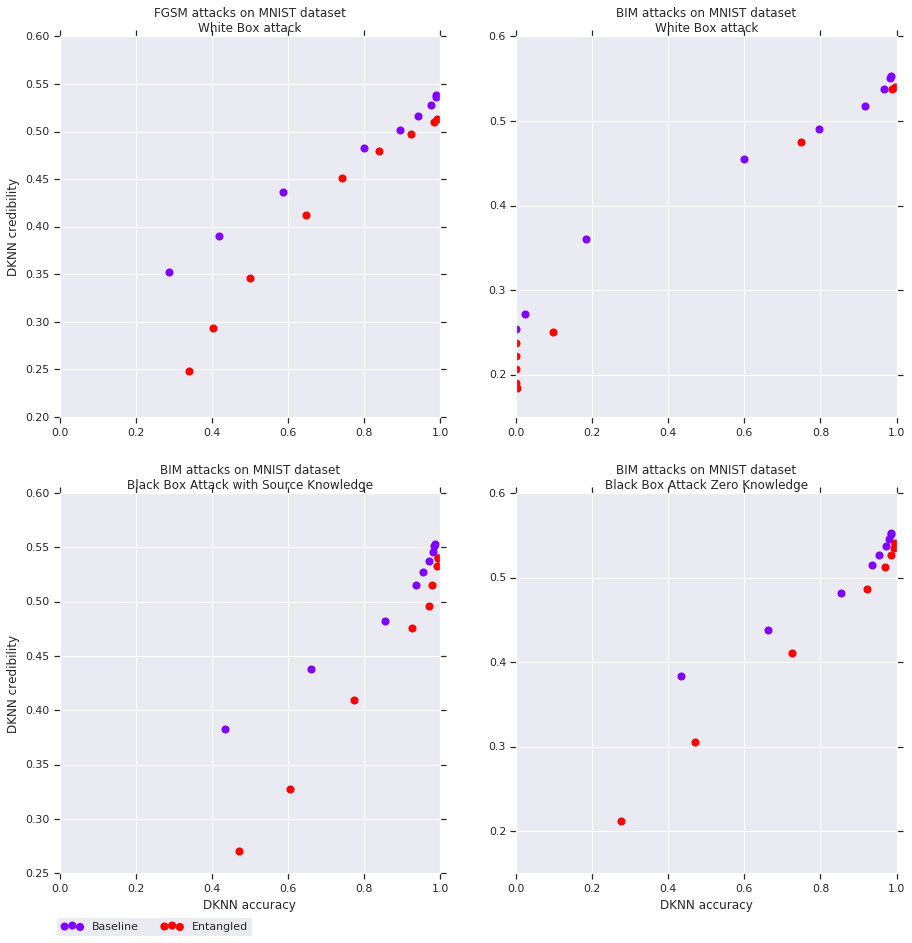}
  \caption{DkNN credibility (prediction support from the training data) as a function of prediction accuracy on MNIST. These plots were created with the same methodology as Figure \ref{fig:dknn_svhn}.}
  \label{fig:dknn_mnist}
\end{figure*}

We include here reports of the DkNN uncertainty calibration on entangled MNIST (Figure~\ref{fig:dknn_mnist}) and FashionMNIST (Figure~\ref{fig:dknn_fashionmnist}) models. The experiment performed is the one described in Section~\ref{ssec:sn-dknn}, where the plot visualizes DkNN credibility as a function of DkNN prediction accuracy.

\begin{figure*}[t]
  \centering
  \includegraphics[width=\linewidth]{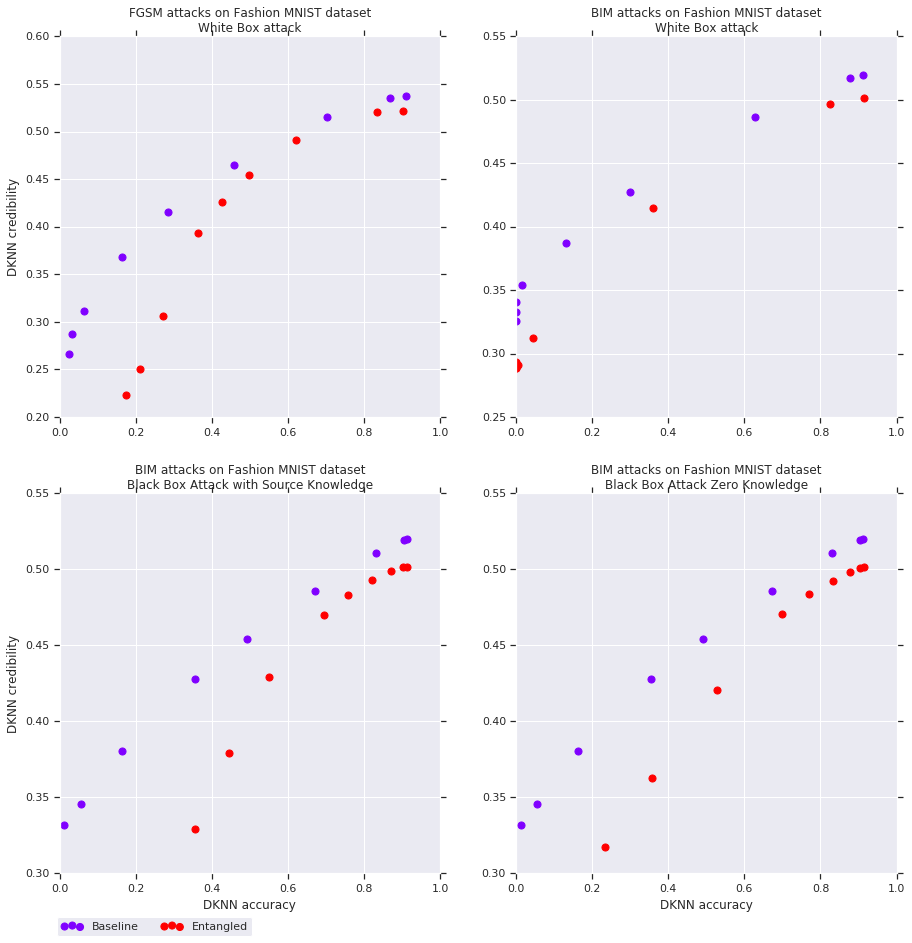}
  \caption{DkNN credibility (prediction support from the training data) as a function of prediction accuracy on FashionMNIST. These plots were created with the same methodology as Figure \ref{fig:dknn_svhn}.}
  \label{fig:dknn_fashionmnist}
\end{figure*}

\begin{figure*}[t]
  \centering
  \includegraphics[width=\linewidth]{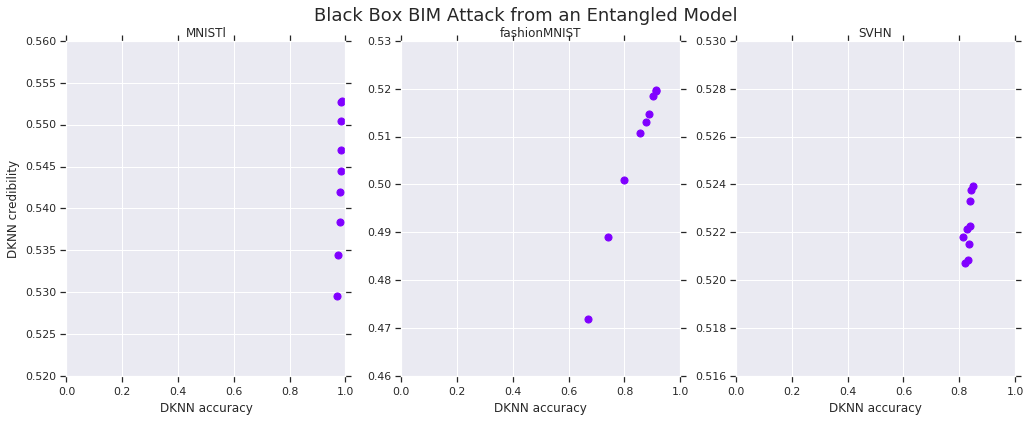}
  \caption{DkNN credibility  as a function of prediction accuracy on all three datasets, where the underlying model is trained with cross-entropy only. Each point corresponds to a set of adversarial examples computed on an entangled model and are transferred to the DkNN with a model trained on cross-entropy only. These plots were created with the same methodology as Figure~\ref{fig:dknn_svhn}. As explained in Section~\ref{ssec:Transferability}, entangled models make poor source models for a black box attack based on transferability: the accuracy of the cross-entropy baseline remains high on all three datasets; the attack is noticeably less effective than in previous settings considered above.}
  \label{fig:dknn_fashionmnist}
\end{figure*}

\section{Out-of-Distribution Test Inputs}
\label{ap:outlier}

This experiment complements results presented with SVHN and CIFAR10 in Section~\ref{ssec:outlier-data}. These results showed that training a network to maximize entanglement  leads to representations that better separate test data from data sampled from a different distribution. We repeat the same experiment on MNIST and notMNIST. 

\begin{figure*}
\includegraphics[width=\linewidth]{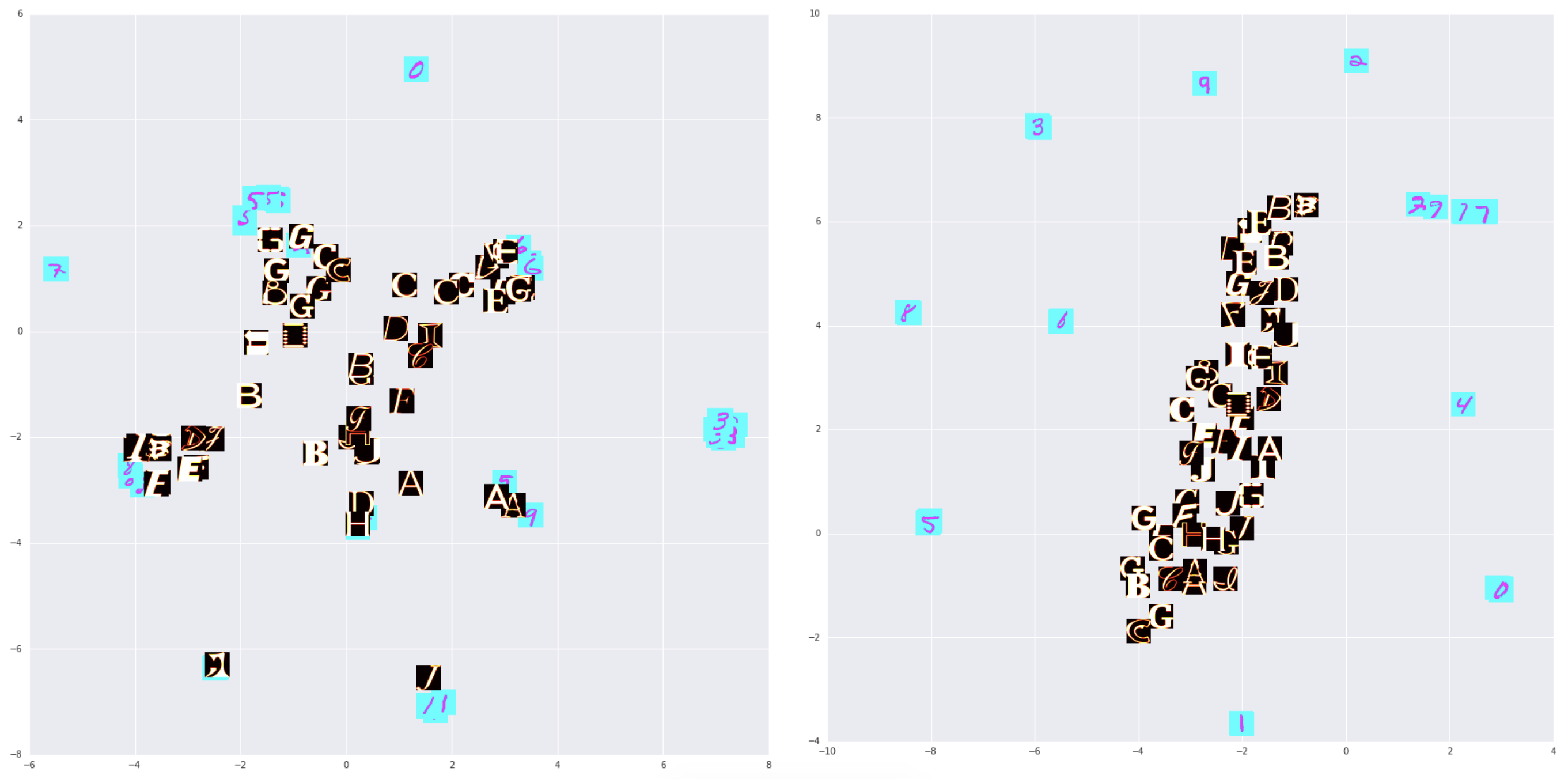}
\caption{t-SNE visualization of representations of in-distribution (MNIST---blue) and out-of-distribution (notMNIST---dark) test data learned by a vanilla (left) and entangled (right) model.}
\label{fig:notmnist}
\end{figure*}

We train a network on MNIST and see what its behavior is like on notMNIST, a data set made up of MNIST-sized typeface characters between letters A and J. Test examples from the notMNIST dataset should thus be projected very differently by a model trained on MNIST, when compared to examples from the MNIST test set. This is indeed what we observe in Figure~\ref{fig:notmnist}, which uses t-SNE to visualize how the logits project MNIST and notMNIST test inputs when a model is trained with cross-entropy only or with the soft nearest neighbor loss to maximize entanglement. We observe that the vanilla model makes confident predictions in the MNIST classes for the notMNIST inputs (because they are projected close to one another), whereas the entangled model separates all of the notMNIST data in a distinct cluster and preserves the MNIST clusters.

\section{Intuition Behind the Improved Calibration of DkNN Uncertainty}
\label{ap:intuition}

\begin{figure*}
\centering
\includegraphics[width=0.8\linewidth]{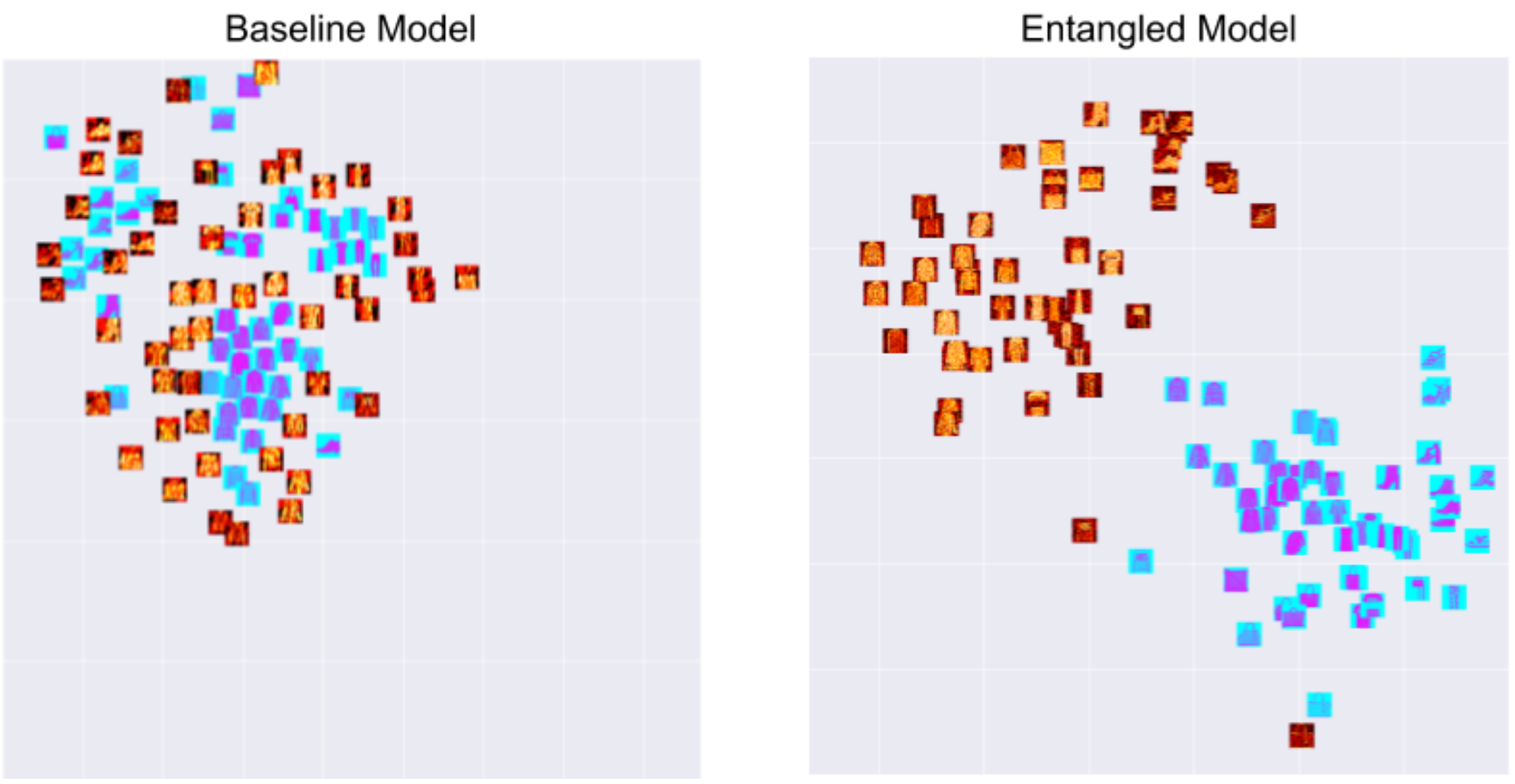}
\caption{t-SNE visualization of the activations from the first hidden layer of a network trained on FashionMNIST. Real data points are plotted in blue whereas adversarial data points are visualized in red. In the vanilla (non-entangled) model, the representations of  adversarial data and real data occupy a similar part of activation space. Instead, in the entangled model, the adversarial data is projected into a separate area. This provide some intuition for why entangled models have better calibrated DkNN uncertainty estimates: it is easier to evaluate support in a particular model  prediction through a nearest neighbor search in the training data given a test point. }
\label{fig:adv-data-projection}
\end{figure*}

In Figure~\ref{fig:adv-data-projection}, we visualize the activations of a hidden layer on real and adversarial test data. In the non-entangled model trained with cross-entropy, the adversarial data is projected close to the real test data. Instead, on the entangled model, the representation separates better the real and adversarial data. This in turn, results in a better estimate of the number of training neighbors that support the prediction made. As a consequence, the DkNN is able to provide more calibrated estimates of uncertainty on entangled representations.

\section{Soft Nearest Neighbor GANs}
\label{ap:gan-sn}

\begin{figure*}
  \centering
  \includegraphics[width=0.7\linewidth]{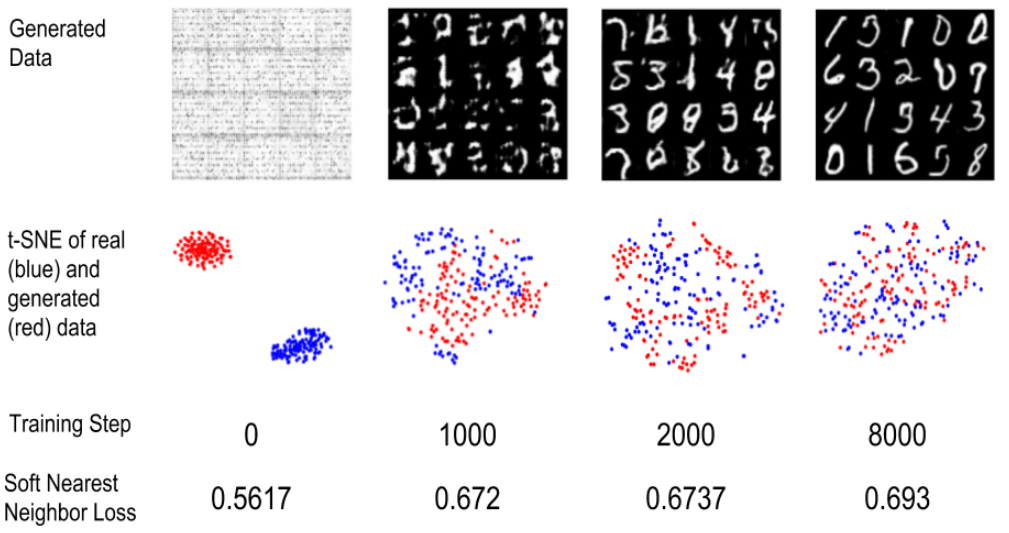}
  \caption{Training progression of a MNIST GAN in which the discriminator minimizes the Soft Nearest Neighbor Loss between real and synthetic data in a learned 10 dimensional space, and the generator maximizes it. We  replaced the output layer of the discriminator with a 10 dimensional vector. This proof-of-concept demonstrates that the soft nearest neighbor loss can be used in a learned space as well as the pixel space, which is explored more thoroughly in the main text.}
  \label{fig:MNIST_learned_space_ent_gan_ent_vis}
\end{figure*}

In Section~\ref{sec:entanglement-metric}, we found that the entanglement loss can effectively replace the discriminator in a GAN setup on MNIST. However, we were unable to scale the same setup to train a CIFAR10 model. We hypothesized that this is due to the fact that the $\ell_2$ distance does not characterize CIFAR10's input domain as well as it does for MNIST. Hence, we run an additional experiment restoring the discriminator but modifying the typical losses used to train GANs: we constrain the generator to entangle synthetic and real data in a 10 dimensional space using the soft nearest neighbor loss while the discriminator is constrained to disentangle the synthetic and real data. While this is simply a proof-of-concept on MNIST, results summarized in Figure~\ref{fig:MNIST_learned_space_ent_gan_ent_vis} demonstrate that this approach deserves further investigation and may scale to larger datasets given the discriminator's ability to learn how to compare points compared to a direct application of the $\ell_2$ distance in the pixel space.

\end{appendices}

\end{document}